\documentclass[11pt]{article}
\usepackage[margin=1in]{geometry}
\usepackage[T1]{fontenc}
\usepackage{lmodern}
\usepackage{amsmath,amssymb,amsthm,mathtools}
\usepackage{microtype}
\usepackage{enumitem}
\usepackage{xcolor}
\usepackage{bbm}
\usepackage{comment}
\usepackage[colorlinks=true,linkcolor=blue!45!black,citecolor=blue!45!black,urlcolor=blue!45!black]{hyperref}
\numberwithin{equation}{section}
\newtheoremstyle{linkedtheorem}{}{}{\itshape}{}{\bfseries}{.}{5pt plus 1pt minus 1pt}
  {\thmname{#1}\thmnumber{ #2}\thmnote{ #3}}
\theoremstyle{linkedtheorem}
\newtheorem{theorem}{Theorem}[section]
\theoremstyle{plain}
\newtheorem{lemma}{Lemma}[section]

\newcommand{\F}{\mathcal F}

\newcommand{\E}{\mathbb E}

\newcommand{\supp}{\operatorname{supp}}

\newcommand{\abs}[1]{\left\lvert #1\right\rvert}
\title{A Minimax Optimal Leave One Out Bound for Bounded Losses}

\author{
Peigan Gao\thanks{The University of Hong Kong. \href{gaopg.daniel@gmail.com}{\texttt{gaopg.daniel@gmail.com}}.}
\and
Jian Qian\thanks{
The University of Hong Kong. \href{mailto:jianqian@hku.hk}{\texttt{jianqian@hku.hk}}.
}
}

\date{November 24, 2026}
\begin{document}
\maketitle

\begin{abstract}

We study proper leave-one-out prediction with \(M\) real-valued experts on \(n\) fixed inputs. Each label is hidden in turn, and the learner uses the remaining labels to choose one of the experts’ predictions at the held-out input. We determine how closely such a rule can match the average loss of the best single expert on the full sample. For any bounded loss that is a continuous nondecreasing function of absolute error and vanishes at zero, let \(B\) denote its maximum value. Over classes of size \(M\) and samples on which the best expert has average loss at most \(r\), the minimax excess loss is of order
\[
\min\{B, \sqrt{\frac{Br\log M}{n}}+\frac{B\log M}{n}.\}
\]
For each class and loss, one deterministic rule attains this rate for every \(r\in[0,B]\). The matching lower bound holds even for randomized proper rules given \(r\). 
\end{abstract}

\section{Introduction}

Suppose that \(M\) experts give their predictions at each of \(n\)
known inputs, with \(n,M\geq 2\). Hide the response at one input and
reveal the other \(n-1\) responses. How accurately can a learner predict the missing
response by choosing one of the values the experts gave at that input?
We measure its loss by hiding each response in turn and averaging the
\(n\) losses. The benchmark is the average loss of the single best
expert, chosen after all \(n\) responses are known. We seek a guarantee
for every labeled sample, with a smaller excess loss on samples where
one expert predicts well.

Lachenbruch and Mickey~\cite{lachenbruch1968} used the deletion of one
observation at a time to estimate misclassification error in discriminant
analysis. Stone~\cite{stone1974} studied how to use such held-out
predictions to choose and assess statistical procedures.
The assumption that the test inputs are known in advance comes from the
transductive formulation of pattern recognition developed by Vapnik and
Chervonenkis~\cite{vapnik1974}. In transductive leave-one-out (LOO)
prediction, the held-out losses are themselves the criterion for
designing the rule.

For binary labels, a single missing label leaves at most two possible
completions. Haussler, Littlestone, and Warmuth introduced the
one-inclusion prediction rule in 1988~\cite{haussler1994}. Its graph
joins two labelings in the class when they differ at just one input.
Orienting an edge specifies which
label to predict, and the out-degree of a labeling counts the mistakes
on that labeling. For a class of VC dimension \(d\), they gave an
orientation with at most \(d\) mistakes, hence LOO error at most \(d/n\),
on every realizable sample. For i.i.d.\ observations, exchangeability
then bounds the expected prediction error after \(n-1\) training
examples by \(d/n\).
Warmuth~\cite{warmuth2004} conjectured in 2004 that the one-inclusion
algorithm would always attain the optimal high-probability risk bound
in realizable PAC learning. Aden-Ali et al.~\cite{adenali2023} refuted
the conjecture by constructing one-inclusion rules with optimal expected
error for which Markov's inequality gives the correct order of the
upper tail.

Long~\cite{long1998} extended the graph orientation to arbitrary binary
labelings, including those outside the class. If the best classifier
makes \(k\) mistakes, his bound gives \(O(k+d)\) LOO mistakes.
Mathiasen, Qian, and Zhivotovskiy~\cite[Theorem~2]{mathiasen2026optimal}
obtained \(k+O(\sqrt{kd}+d)\) mistakes and used this fixed-sample
inequality to derive an optimal agnostic PAC bound. When \(k\) is
large compared with \(d\), the coefficient one on \(k\) means that
only \(O(\sqrt{kd})\) additional mistakes are allowed.

For real-valued responses, the cost of an error depends on its size.
Bartlett and Long~\cite{bartlett1998} bounded the expected absolute
prediction error for classes of real-valued functions in terms of a
scale-sensitive dimension. Forster and Warmuth~\cite{forster2002}
used LOO bounds to control expected prediction loss in linear regression
and density estimation. More directly, Qian and
Xu~\cite[Corollary~10]{qian2026multiplicative} obtained a fixed-sample
bound for a finite class by averaging predictions within empirical-loss
level sets on each LOO sample and taking a median across levels. For a
loss bounded by
\(B\), convex in the prediction and nondecreasing in the distance to the
response, their bound is eight times the best average empirical loss plus
\(104B(\log M)/n\). At positive comparator loss, the factor eight leaves
room for an excess proportional to that loss even as \(n\) grows. H{\o}gsgaard, Rebeschini, and Wegel~\cite[Proposition~3]{hogsgaard2026aggregation}
obtained a deterministic bound with coefficient one for the exponentially
weighted LOO predictor. Suppose \(0\leq\ell\leq B\) and, for every response
\(y\), the function \(\ell(\cdot,y)\) is \(L\)-Lipschitz and
\(\mu\)-strongly convex. If the temperature \(T>0\) satisfies
\((L^2/T)\exp(B/T)\leq\mu/2\), then on every fixed sample
\[
  \frac{1}{n}\sum_{i=1}^n
    \ell\!\left(\sum_{a=1}^M p_a^{(-i)}f^{(a)}(i),y_i\right)
  \leq
  \min_{1\leq a\leq M}\frac{1}{n}\sum_{i=1}^n
    \ell(f^{(a)}(i),y_i)
  +\frac{T\log M}{n}.
\]
Here \(f^{(a)}(i)\) is expert \(a\)'s prediction at input \(i\).
Its weight \(p_a^{(-i)}\) is proportional to
\(\exp\{-T^{-1}\sum_{j\ne i}\ell(f^{(a)}(j),y_j)\}\), with the
weights normalized to sum to one over the \(M\) experts. For squared
loss on \([0,1]\), they show directly that \(T=4\) suffices.

In statistical model selection aggregation, a learner may average
dictionary predictions while competing with the best single member.
Tsybakov~\cite{tsybakov2003} identified the \((\log M)/n\) benchmark
under squared loss. Lecu\'e and Mendelson~\cite{lecue2013}
showed that exponential weighting can miss this rate at low temperatures
and asked whether a sufficiently large fixed temperature attains it in
expectation.
H{\o}gsgaard, Rebeschini, and Wegel~\cite[Theorem~1]{hogsgaard2026aggregation}
answered this question by applying their deterministic LOO inequality
to \(n+1\) i.i.d.\ observations. Each held-out observation is independent
of the other \(n\), so the expected LOO loss equals the expected
population risk of the aggregate trained on \(n\) observations.
Since the expected empirical minimum is at most the smallest population
risk among the experts, the resulting expected excess risk is at most
\(T(\log M)/(n+1)\). If two experts predict \(0\) and \(1\), their
aggregate predicts \(1/2\) when the weights are equal. A proper LOO
rule must choose one of the two expert values using the remaining labels.

We ask whether the coefficient-one bound from binary prediction extends
to this choice among real-valued expert predictions, even for nonconvex
losses. Let \(\mathcal F\subseteq[0,1]^{[n]}\) be the trace of the
\(M\) experts and write \(S=((i,y_i))_{i=1}^n\), with
\(y_i\in[0,1]\). The loss is
\(\ell_\psi(a,y)=\psi(|a-y|)\), where
\(\psi:[0,1]\to[0,1]\) is continuous and nondecreasing,
\(\psi(0)=0\), and \(B=\psi(1)>0\). Write \(y_{-i}\) for the labels
observed when \(y_i\) is hidden. The two averages being compared are
\[
  \widehat L^{\mathrm{LOO}}_{S,\psi}(A)
  =\frac{1}{n}\sum_{i=1}^n\ell_\psi(A_i(y_{-i}),y_i),
  \qquad
  \widehat L^*_{S,\psi}
  =\frac{1}{n}\min_{f\in\mathcal F}\sum_{i=1}^n\ell_\psi(f(i),y_i).
\]
We prove that, for each trace and loss, one can fix
deterministic Borel maps \(A_i(y_{-i})\in\{f(i):f\in\mathcal F\}\)
such that every labeling satisfies
\[
  \widehat L^{\mathrm{LOO}}_{S,\psi}(A)-\widehat L^*_{S,\psi}
  \leq 2^{15}\left(
    \sqrt{\frac{B\widehat L^*_{S,\psi}\log M}{n}}
    +\frac{B\log M}{n}\right).
\]
If one expert incurs zero loss, only \(2^{15}B(\log M)/n\) remains.
Once \(\widehat L^*_{S,\psi}\geq B(\log M)/n\), the square-root term
dominates. The same rule covers both regimes without knowing the
comparator loss in advance. The bound covers absolute and
squared loss as well as the nonconvex capped absolute loss
\(\psi(t)=\min\{t,\tau\}\), for which \(B=\tau\) for every \(0<\tau<1\).

The square-root dependence on the best expert's loss also occurs in
sequential prediction with expert advice~\cite{freund1997,deRooij}.
Exponential weights control a weighted average of the experts' losses.
Sampling an expert with those weights realizes that loss in
expectation. A deterministic rule must coordinate its choices across
the held-out inputs to control the sum of the losses it actually incurs.
The rounding lemma in Section~\ref{sec:rounding} uses the solution of
the Koml\'os conjecture by Guo, Fang, and Lu~\cite{guo2026}, in the
form with signing constant \(36\) proved by Karingula and
Lovett~\cite[Theorem~1.2]{KL}.

\section{Problem setup}

\subsection{Settings and notations}
Let $n,M\ge2$ and fix a nonempty finite trace
\[
\mathcal F=\{f^{(1)},\ldots,f^{(M)}\}\subseteq[0,1]^{[n]}.
\]
Thus $f^{(a)}(i)$ is expert $a$'s prediction at coordinate $i$.
For arbitrary labels $y_1,\ldots,y_n\in[0,1]$, write
$v_y=(y_1,\ldots,y_n)$ and $S=((i,y_i))_{i=1}^n$.
The coordinates themselves serve as the fixed inputs; a trace on any
other fixed input sequence gives the same problem. Write $S^{-i}$
for the sample with observation $i$ removed, and
$y_{-i}=(y_j)_{j\ne i}$ for its observed labels.

An admissible $\psi:[0,1]\to[0,1]$ is continuous and nondecreasing,
with $\psi(0)=0$ and $B=\psi(1)>0$. Its two-argument loss is
$\ell_\psi(a,y)=\psi(|a-y|)$. A deterministic proper leave-one-out
rule consists of Borel prediction maps
\[
\widehat y_i=A_i(y_{-i})\in\{f^{(1)}(i),\ldots,f^{(M)}(i)\},
\qquad i\in[n].
\]
which means it is included by results that the experts give.
The rule may use the entire fixed trace and the labels in $S^{-i}$,
but not $y_i$ or the complete-sample optimum. Define
\[
\widehat L_{S,\psi}^{*}
=\frac1n\min_{f\in\mathcal F}\sum_{i=1}^n\ell_\psi(f(i),y_i),
\]
\[
\widehat L_{S,\psi}^{\mathrm{LOO}}(A)
=\frac1n\sum_{i=1}^n\ell_\psi(\widehat y_i,y_i),
\]
Here, $\widehat{L_{S,\psi}^{*}}$ reflects the best result given by experts, $\widehat L_{S,\psi}^{\mathrm{LOO}}(A)$ shows the loss of the prediction when given rule $A$. We omit $A$ from $\widehat L^{\mathrm{LOO}}$ when the rule is fixed. All logarithms are natural unless a base is displayed.

An interval $I$, i.e., a set of consecutive sample indices of the form
$I=\{b+1,\ldots,k\}$, where $0\le b<k\le n$. Since $\psi(1)=B$, write the normalized loss and the best total loss on an interval \(I\subseteq[n]\) as
\begin{equation}
 \ell_{i,a}(t)=\frac{\ell_\psi(f^{(a)}(i),t)}{B},\qquad
 L_I^*=\min_{a\in[M]}\sum_{i\in I}\ell_{i,a}(y_i),\qquad
 L_n^*=L_{[n]}^*=\frac nB\widehat L^*_{S,\psi}.
 \label{eq:normalization}
\end{equation}

The first result is the finite signing statement in~\cite[Theorem~1.2]{KL}, which is the main tool for building the results. It gives a $l^{\infty}$ bound of sums of signs,
\begin{theorem}\label{thm:signing}
For every finite collection $u_1,\ldots,u_N\in\mathbb R^d$ with
$\|u_j\|_2\le1$, there are signs $\varepsilon_j\in\{-1,1\}$ such that
\[
\left\|\sum_{j=1}^N\varepsilon_j u_j\right\|_\infty\le36.
\]
\end{theorem}

\subsection{Main results}

Our first main theorem gives a general admissible losses for function $\psi$ by applying Theorem \ref{thm:signing}.

\begin{theorem}[(General admissible losses)]\label{thm:GEN}

For every such trace and loss, there exist deterministic Borel maps
\(a_i:[0,1]^{n-1}\to[M]\) such that the rule \(A\), defined by
\(A_i(y_{-i})=f^{(a_i(y_{-i}))}(i)\), satisfies the following
inequality for every sample \(S\),
\begin{equation}
 \widehat L^{\mathrm{LOO}}_{S,\psi}(A)
 \le \widehat L^*_{S,\psi}
 +2^{15}\left(
     \sqrt{\frac{B\widehat L^*_{S,\psi}\log M}{n}}
        +\frac{B\log M}{n}\right).
 \label{eq:oracle}
\end{equation}
The maps depend only on \(n,\F,\psi\).

\end{theorem}

Next theorem shows that our result is minimax optimal up to constants.

\begin{theorem}[(Minimax Optimality up to universal constants)]\label{minimax}

Fix $n,M\ge2$ and an admissible $\psi$, with $B=\psi(1)>0$.
For every $0\le r\le B$,
\[
\begin{aligned}
 &\sup_{\substack{\mathcal F\subset[0,1]^{[n]}\\ |\mathcal F|=M}}
   \ \inf_{A\ \mathrm{deterministic,\ proper}}
   \ \sup_{\substack{v_y\in[0,1]^n\\
                      \widehat L^*_{S,\psi}\le r}}
   \left(\widehat L^{\mathrm{LOO}}_{S,\psi}(A)
                  -\widehat L^*_{S,\psi}\right)
 \asymp
    \min\!\left\{B,\
      \sqrt{\frac{B r\log M}{n}}+\frac{B\log M}{n}\right\}.
\end{aligned}
\]
The constants are universal. The infimum is over the Borel prediction
maps of Section~2.1 for the given trace $\mathcal F$. The upper bound
is attained up to a universal factor by one rule for all $r$. The
lower bound continues to hold if the rule knows $r$ and may randomize,
with its loss then taken in expectation on each fixed sample.

\end{theorem}

\subsection{Proof architecture}

In section 3, we discuss the learning algorithm of exponential weights and prove the guarantee. In section 4, we will first show the supporting lemmas for the complete proof, where we apply Theorem \ref{thm:signing} to prove the grouped rounding lemma. In section 5, we will present the full proof of the main theorem by separating the small scale and the large scale. In section 6, we will give some further discussions including the minimax optimality up to constants.

\section{Exponential weights via learning}

Considering that there are $M$ different experts, we apply the Hedge weight calculation \cite[Section~2.1]{deRooij} to help renew the weights on experts with a fixed learning rate $\beta$. The following lemma emphasizes that we can control the cost by a fixed surviving expert.

\begin{lemma}\label{lem:mass}
On an interval \(I\), start each of \(M\) experts with weight \(1/M\).
Before step \(i\), let \(p_{i,a}\) be the current weight on expert $\textbf{a}$ divided by
the total remaining weight, where $\sum_{a} p _{i,a}=1$. Give a retained expert $\textbf{a}$'s cost in round $i$, \(c_{i,a}\in[0,1]\). Write $h_i=\sum_{a\ \mathrm{present\ at\ the\ start\ of\ round }\ i} p_{i,a}c_{i,a}$ be the weighted average loss over the present experts in round $i$.
Then multiply its weight $W_{i-1}p_{i,a}$ by \(e^{-\beta c_{i,a}}\), and optionally delete
experts, leaving at least one. For \(0<\beta\le1\) and each fixed expert \(a^*\)
surviving the whole interval,
\begin{equation}
 \sum_{i\in I}\ h_i
 \le \frac{\beta}{1-e^{-\beta}}
        \left(\sum_{i\in I}c_{i,a^*}+\frac{\log M}{\beta}\right)
 \leq (1+\beta)\left(\sum_{i\in I}c_{i,a^*}+\frac{\log M}{\beta}\right) .
 \label{eq:mass}
\end{equation}
\end{lemma}
\begin{proof}
    Let the interval be $I=\{b+1,b+2,\cdots,k\}$, and \(W_i\) be the total weight after the update and deletions at \(i\). At the beginning $W_b=M\cdot \frac{1}{M}=1.$ Starting at round $i$, expert $\textbf{a}$ has weights $W_{i-1}p_{i,a}$, the renewal action in round $i$ makes the weights become $W_{i-1}p_{i,a}e^{-\beta c_{i,a}}$.  

    Note that, by convexity of \(u\mapsto e^{-\beta u}\), for every
    \(u\in[0,1]\),
    \[
    e^{-\beta u}\leq (1-u)+ue^{-\beta}
    =1-(1-e^{-\beta})u.
    \]
    Applying this bound with \(u=c_{i,a}\) gives
    \[
    \begin{aligned}
    \frac{W_i}{W_{i-1}}
    &\leq
    \sum_{\substack{\text{experts }a\text{ present}\\
                    \text{at the start of round }i}}
    p_{i,a}e^{-\beta c_{i,a}}
    \leq
    \sum_{\substack{\text{experts }a\text{ present}\\
                    \text{at the start of round }i}}
    p_{i,a}\bigl[1-(1-e^{-\beta})c_{i,a}\bigr] \\
    &= 1-(1-e^{-\beta})h_i
    \leq e^{-(1-e^{-\beta})h_i}.
    \end{aligned}
    \]
    The equality follows from the definition of \(h_i\) and the fact that
    the normalized weights of the experts present at the start of round
    \(i\) sum to one. The last inequality uses \(1-x\leq e^{-x}\) with
    \(x=(1-e^{-\beta})h_i\). 
    Consequently, 
    \[
    W_k=W_b \cdot \prod_{i=b+1}^{k} \frac{W_i}{W_{i-1} } \leq \exp\{-(1-e^{-\beta})\sum_{i\in I} h_i\}.
    \]

    For a deterministic expert $a^{*}$, its final weight can be calculated as $\frac{1}{M}\exp\{-\beta \sum_{i \in I} c_{i,a^{*}}\}$. Since $a^*$ survives through the interval $I$, the final weights must include weights from $a^{*}$, which gives
    \[
    \frac{1}{M}\exp\{-\beta \sum_{i \in I} c_{i,a^{*}}\} \leq W_k \leq \exp\{-(1-e^{-\beta})\sum_{i\in I} h_i\}.
    \]
    Taking logarithms gives,
    \[
    \sum_{i\in I}\ h_i
     \le \frac{\beta}{1-e^{-\beta}}
            \left(\sum_{i\in I}c_{i,a^*}+\frac{\log M}{\beta}\right)
     \leq (1+\beta)\left(\sum_{i\in I}c_{i,a^*}+\frac{\log M}{\beta}\right) .
    \]
    The last inequality uses $e^{-\beta} \leq \frac{1}{1+\beta}$ to prove $\frac{\beta}{1-e^{-\beta}}\leq 1+\beta$.
\end{proof}

\section{Support-preserving grouped rounding}\label{sec:rounding}

Suppose that the group is a setting which needs to take action, where actions per group means choosing expert $A$. The row of the table is a complete data set $z$ which is used to test the performance of the action table. Let $c_{zga}$ means the cost contributed to row $z$ by choosing action $a$ in group $g$. This forms a table, where the columes are different groups together with actions. The next lemma shows that there exists choices good enough.

\begin{lemma}[Countable grouped rounding]\label{lem:countable-rounding}
Let the row and group sets be countable, with finitely many actions
per group and finitely many contributing groups per row, i.e. for each row $z$, only finite many groups $g$ have $c_{zga} \neq 0$ for some action $a$. Let $p_g$
be a probability vector in group $g$. Suppose every within-group
action-difference column satisfies
\[
\sum_z(c_{zga}-c_{zgb})^2\le D_1^2.
\]
where $D_1 \geq 0$ is a constant. Then, there are choices $a_g\in\supp(p_g)$ such that
\[
\left|\sum_g c_{z,g,a_g}-\sum_{g,a}p_{ga}c_{zga}\right|
\le36D_1
\]
on every row. Initially pure groups remain fixed.
\end{lemma}

\begin{proof}
If $D_1=0$, all actions in each group have the same row coefficients,
so any supported choices work. 

Assume $D_1>0$, we first restrict to finitely many rows and the probability vectors are dyadic rational; only finitely many groups occur.
Suppose dyadic probabilities with denominator $2^K$ for the probability vector at the beginning. Consider the current probability vector be $(x_{ga})$, where the original probability is $(p_{ga})$ and the dyadic probabilities with denominator $2^k$. Pair the odd
numerators inside each group at denominator $2^k$. For a pair (a,b), renewal the probability by:
\[x_{ga} \leftarrow x_{ga}+\epsilon_{gab}2^{-k}, \quad x_{gb} \leftarrow x_{gb}-\epsilon_{gab}2^{-k}, \quad \space \epsilon_{gab} \in \{-1,+1\}.\]
Here $\epsilon_{gab}$ will be chosen by applying Theorem \ref{thm:signing}
to all paired difference columns divided by $D_1$, there exists $\epsilon_{gab}$ such that,
\[
\abs{\sum_{\text{every paired (g,a,b)}}\epsilon_{gab}(c_{zga}-c_{zgb})} \leq 36D_1, \quad \forall z.
\]
The change of the weighted sum on each row in renewal round for denominator $2^k$ satisfies:
\[
\abs{\sum_{\text{every paired (g,a,b)}}\epsilon_{gab}2^{-k}(c_{zga}-c_{zgb})} \leq 36 \cdot 2^{-k} D_1, \quad \forall z.
\]
Numerators become even and remain nonnegative. After this round, the denominator decrease to $2^{k-1}$.
Repeating for $k=K,\ldots,1$ produces a pure assignment with error less than $36D_1$ and preserves the original
supports, since
\[
\left|\sum_g c_{z,g,a_g}-\sum_{g,a}p_{ga}c_{zga}\right|
\le \sum_{k=1}^{K} 36 \cdot 2^{-k} D_1 <36D_1.
\]
Thus we finish the proof for finite dyadic probabilities. 

Approximate arbitrary probabilities by dyadic probabilities on their
original supports. There are finitely many assignments to the groups
under consideration, so one assignment recurs along an infinite
subsequence. Taking the limit gives discrepancy at most $36D_1$ on these
rows. This proves the finite-row assertion.

For countably many rows, the product \(\prod_g\supp(p_g)\) is
compact. Each row inequality is closed because it involves only
finitely many groups, and the finite-row result gives the finite
intersection property. An assignment in the intersection proves
the claim.
\end{proof}

\section{Proof of generalized inequality}

Recall the notation that:

\begin{equation}
 \ell_{i,a}(t)=\frac{\ell_\psi(f^{(a)}(i),t)}{B} \in [0,1],\qquad
 L_I^*=\min_{a\in[M]}\sum_{i\in I}\ell_{i,a}(y_i),\qquad
 L_n^*=L_{[n]}^*=\frac nB\widehat L^*_{S,\psi}.
 \label{eq:normalization-recall}
\end{equation}

Recall the model of Lemma \ref{lem:mass}.
The first interval starts at coordinate \(1\), with each of the \(M\)
experts assigned weight \(1/M\) and cumulative loss zero. We set the
loss threshold for this interval to \(R=\log(2M)\). Both \(R\) and the
learning rate \(\beta=\sqrt{(\log M)/R}\) remain fixed throughout the
interval. The initial choice \(R=\log(2M)>\log M\) ensures that
\(\beta\le 1\), as required by Lemma~\ref{lem:mass}. Since \(\log(2M)<M^8\) at the beginning, the first interval uses the small-scale stopping
rule, where the small-scale means the $R<M^{8}$ case. After observing \(y_k\), we compute the smallest cumulative loss
of a single expert on the first \(k\) coordinates,
\[
L^*_{\{1,\ldots,k\}}
=\min_{a\in[M]}\sum_{i=1}^{k}\ell_{i,a}(y_i).
\]

The first interval ends at the first coordinate \(k\) for which
\(L^*_{\{1,\ldots,k\}}>R\), and it includes that coordinate.
If the loss never exceeds \(R\), the first interval is all of
\(\{1,\ldots,n\}\). If the threshold is first crossed at \(k<n\), the
next interval starts at coordinate \(k+1\). At this and every subsequent
restart, we reset all scores to zero, multiply the current threshold
\(R\) by four, and recompute \(\beta\) using the new value of \(R\).

In this section, we will show for every interval $I$ with threshold $R$, its deterministic
prediction loss satisfies
\[
\frac{1}{B}\sum_{i\in I} \ell_{\psi}(\widehat{y_i},y_i) \leq L_{I}^{*}+2^{12} \sqrt{R \log M}.
\]
If another interval follows $I$, then
\[
L_{I}^{*} > \frac{R}{2}.
\]

\subsection{Small scale}

Suppose \(R<M^8\) and the interval starts after coordinate \(b\), i.e. we can write the interval as \(I=\{b+1,\ldots,k\}\). Before predicting \(i\),
assign expert \(a\) the weight
\[
 p_{i,a}=\frac{\exp(-\beta\sum_{j=b+1}^{i-1}\ell_{j,a}(y_j))}
                {\sum_{a'=1}^M\exp(-\beta\sum_{j=b+1}^{i-1}\ell_{j,a'}(y_j))}.
\]
Note that the sum ends at $i-1$, computing $p_{i,a}$ only needs the labels before $i$. After dealing with coordinate $k$, if 
\[
L_{\{b+1,\cdots,k\}}^*=\min_{a\in[M]}\sum_{j=b+1}^{k}\ell_{j,a}(y_j)>R,
\]
then we will end this interval by definition. Since all experts' losses are controlled by $[0,1]$, this gives
\[
L_{\{b+1,\cdots,k-1\}}^* \leq L_{\{b+1,\cdots,k\}}^* \leq L_{\{b+1,\cdots,k-1\}}^*+1 \leq R+1.
\]

The following construction coordinates the predictions using the
remaining labels. Fix the modulus
\[
       K=4M^9,\qquad \log K\le11\log M .
\]
The modulus is the same for every small scale. We first give the
one-coordinate construction for any distribution on the residues,
then specify that distribution using the labels before \(i\).

\paragraph{A choice for one coordinate.}
Fix \(i\). At this step the expert weights \(p_{i,a}\) and a probability
law \(\mu\) on \(\{0,\ldots,K-1\}\) are given. For a possible label \(t\)
write \(h_i(t)=\sum_a p_{i,a}\ell_{i,a}(t)\). We construct one map
\(\phi_i\) that works for every \(t\).

Draw an expert with probabilities \(p_{i,a}\) and let \(X\) be its
prediction. Its mean normalized loss is exactly
\[
 \E\!\left[\frac{\ell_\psi(X,t)}B\right]
 =\sum_a p_{i,a}\ell_{i,a}(t)=h_i(t).
\]
What we want is a deterministic prediction, and this is the purpose of the following construction.

First, we will construct $x_{med}$. Let \(x_{\mathrm{med}}\) be a weighted median of \(X\), i.e. it is the least prediction with \(\mathbb{P}(X\le x_{\mathrm{med}})\ge1/2\). Then
\(\mathbb{P}(X\ge x_{\mathrm{med}})\ge1/2\) as well. For any \(t\ge x_{\mathrm{med}}\),
each \(X\le x_{\mathrm{med}}\) has $X\le x_{\mathrm{med}} \le t$, which gives $\abs{X-t}\geq \abs{x_{med}-t}$. By monotonicity of function $\psi$, 
\[
h_i(t) = \E\!\left[\frac{\ell_\psi(X,t)}B\right] \geq \mathbb{P}(X \leq x_{med}) \cdot \frac{\ell_\psi(x_{med},t)}B \geq \frac{1}{2} \frac{\ell_\psi(x_{med},t)}B.
\]
For \(t<x_{\mathrm{med}}\), the same is true for each \(X\ge x_{\mathrm{med}}\). Thus in either
case
\[
 \frac{\ell_\psi(x_{\mathrm{med}},t)}B\le 2h_i(t).
\]

Then, write \(x_1<\cdots<x_m\) for the distinct predictions in the trace at
coordinate \(i\). Since every weight \(p_{i,a}\) is positive, these
are precisely the possible values of \(X\). Let
\(t_j=(x_j+x_{j+1})/2\) for \(1\le j<m\), and put
\[
 \tau_i(t)=\#\{j<m:t_j<t\},\qquad \tau_i(t_j)=j-1 .
\]
These counts depend only on the trace and on \(t\), and
\(0\le\tau_i(t)<m\le M\). Fix this definition of \(\tau_i\) for
every coordinate of the sample. A small-scale predictor can then
use the counts of all other coordinates, regardless of their eventual scales.

Write \([v]_K\) for the residue in \(\{0,\ldots,K-1\}\), i.e. $[v]_K \equiv v \mod K$. $\mu$ be the probability over the residue in \(\{0,\ldots,K-1\}\), and take $r$ a random residue variable with $r \sim \mu$. 

Arrange the residues \(0,\ldots,K-1\) in cyclic order. Choosing an
\emph{origin} means choosing a residue \(o\) to serve as position \(0\). We wish to choose \(o\)
so that a draw \(r\sim\mu\) has small probability of falling in the
first \(M\) positions \(0,\ldots,M-1\) or the last \(M\) positions
\(K-M,\ldots,K-1\). Fix a residue \(v\), as \(o\) runs through \(0,\ldots,K-1\), the position
\([v-o]_K\) takes each value in \(\{0,\ldots,K-1\}\) exactly once.
Hence exactly \(M\) choices of \(o\) place \(v\) in the first \(M\)
positions, and exactly \(M\) choices place it in the last \(M\)
positions. These two sets of choices are disjoint since \(K\ge 2M\).
Therefore 
\[
\begin{aligned}
&\frac{1}{K}\sum_{o=0}^{K-1}
  \mathbb{P}_{r\sim\mu}\bigl([r-o]_K<M
                  \ \text{or}\ [r-o]_K\ge K-M\bigr) \\
&\qquad =
  \sum_{v=0}^{K-1}\mu(v)\,\frac{1}{K}
  \sum_{o=0}^{K-1}
  \mathbf{1}\bigl\{[v-o]_K<M
                  \ \text{or}\ [v-o]_K\ge K-M\bigr\} \\
&\qquad =
  \sum_{v=0}^{K-1}\mu(v)\frac{2M}{K}
  =\frac{2M}{K}.
\end{aligned}
\]
This ensures that at least one origin
has probability at most \(2M/K\) of placing \(r\) in the first or last
\(M\) positions. Choose the smallest such \(o\) in the order
\(0,1,\ldots,K-1\), and set \(s=[r-o]_K\). In the next step we regard
\(s\) as an ordinary integer and compute \(s-\tau_i(t)\) in
\(\mathbb{Z}\).

For each \(j<m\) take \(b_j\) to be the largest
integer \(b_j\in\{0,\ldots,K-1\}\) with
\(\mathbb{P}(s\ge b_j)\ge\mathbb{P}(X\le x_j)\). By this choice,
\[
 \mathbb{P}(s\ge b_j+1)<\mathbb{P}(X\le x_j)\le
 \mathbb{P}(s\ge b_j).
\]
Define a prediction \(x(v) \in \{x_1,\cdots,x_m\}\) for each integer \(v\) by
\[
 x(v)\le x_j\quad\Longleftrightarrow\quad
                v\ge b_j-j+1,\qquad 1\le j<m .
\]
The thresholds on the right decrease with \(j\). Indeed, the probabilities
\(\mathbb{P}(X\le x_j)\) increase, so \(b_j\) decreases weakly, while
\(j\) increases. Thus the comparisons determine one of
\(x_1,\ldots,x_m\). When \(m=1\), take \(x(v)=x_1\).

Fix a possible label \(t\). The nearest prediction is \(x_{\tau_i(t)+1}\),
taking the smaller site when \(t\) is a midpoint.
Write \(Y=x(s-\tau_i(t))\) for the prediction on the integer line.
The definition of \(x(v)\) gives
\[
 \mathbb{P}(Y\le x_j)=\mathbb{P}(s-\tau_{i}(t) \geq b_j-j+1)=\mathbb{P}(s\ge b_j+\tau_i(t)+1-j).
\]
To the left of the nearest site \(j<\tau_i(t)+1\), the threshold is at least
\(b_j+1\). For \(j\ge \tau_i(t)+1\), it is at most \(b_j\). Thus the left and
right tails away from \(x_{\tau_i(t)+1}\) satisfy
\begin{align*}
 j<\tau_i(t)+1:\quad \mathbb{P}(Y\le x_j)
 &\le\mathbb{P}(s\ge b_j+1)\le\mathbb{P}(X\le x_j),\\
 \tau_i(t)+1\le j<m:\quad \mathbb{P}(Y>x_j)
 &=1-\mathbb{P}(s\ge b_j+\tau_i(t)+1-j)
 \le1-\mathbb{P}(s\ge b_j)\le\mathbb{P}(X>x_j).
\end{align*}
The distances \(|x_j-t|\) decrease down to their minimum at \(\tau_i(t)+1\)
and then increase, with a possible tie at the minimum. Since
\(\psi\) is nondecreasing, the sites whose normalized loss exceeds
a level \(v\) therefore form either the full set or a disjoint union
of a left tail and a right tail excluding \(x_{\tau_i(t)+1}\). Either tail may
be empty. Adding the two displayed probability bounds gives
\[
 \mathbb{P}\{\ell_\psi(Y,t)>Bv\}
 \le\mathbb{P}\{\ell_\psi(X,t)>Bv\},\qquad 0\le v\le1 .
\]
For the full set both probabilities are one. Integrating over \(v\) to calculate the expectation therefore gives
\[
 \E\!\left[\frac{\ell_\psi(Y,t)}B\right]
 =\int_0^1\mathbb{P}\{\ell_\psi(Y,t)>Bv\}\,dv
 \le\int_0^1\mathbb{P}\{\ell_\psi(X,t)>Bv\}\,dv=h_i(t).
\]

Return to residues by setting
\[
 \phi_i(v)=
 \begin{cases}
   x([v-o]_K),&[v-o]_K<K-M,\\
   x_{\mathrm{med}},&K-M\le[v-o]_K<K .
 \end{cases}
\]
To compare \(\phi_i(r-\tau_i(t))\) with \(x(s-\tau_i(t))\), consider
first \(s<\tau_i(t)\). Then \(s<M\), and reduction modulo \(K\) sends
\(s-\tau_i(t)\) into the last \(M\) residues. Hence the cyclic rule
uses \(x_{\mathrm{med}}\). If \(s\ge\tau_i(t)\), no reduction is needed. The cyclic
rule agrees with the line rule except when \(s-\tau_i(t)\ge K-M\).
In that case \(s\ge K-M\) and it again uses \(x_{\mathrm{med}}\). Thus on every
disagreement the cyclic rule uses the median, and the probability of
disagreement is at most \(2M/K\). For
\(\ell_\psi(\phi_i(r-\tau_i(t)),t)/B\) 
we obtain
\begin{align*}
\frac{\ell_\psi(\phi_i(r-\tau_i(t)),t)}B
&=
\mathbf{1}_{\{\phi_i(r-\tau_i(t))=Y\}}
   \frac{\ell_\psi(Y,t)}{B}
+
\mathbf{1}_{\{\phi_i(r-\tau_i(t))\ne Y\}}
   \frac{\ell_\psi(x_{\mathrm{med}},t)}{B} \\
&\le
\frac{\ell_\psi(Y,t)}{B}
+
\mathbf{1}_{\{\phi_i(r-\tau_i(t))\ne Y\}}
   \frac{\ell_\psi(x_{\mathrm{med}},t)}{B},
\end{align*}
where the inequality follows from \(\ell_\psi(Y,t)\ge 0\).
Since \(t\) and \(x_{\mathrm{med}}\) are fixed as \(r\) varies, taking
expectations gives
\begin{align*}
\E_{r\sim\mu}[\frac{\ell_\psi(\phi_i(r-\tau_i(t)),t)}B]
&\le
\E_{r\sim\mu}\!\left[\frac{\ell_\psi(Y,t)}{B}\right]
+
\mathbb{P}_{r\sim\mu}\bigl(\phi_i(r-\tau_i(t))\ne Y\bigr)
   \frac{\ell_\psi(x_{\mathrm{med}},t)}{B} \\
&\le
\E_{r\sim\mu}\!\left[\frac{\ell_\psi(Y,t)}{B}\right]
+\frac{2M}{K}\frac{\ell_\psi(x_{\mathrm{med}},t)}{B} \\
&\le h_i(t) + \frac{2M}{K} \cdot 2h_i(t)=\left(1+\frac{4M}{K}\right)h_i(t)\\
&=(1+M^{-8})h_i(t).
\end{align*}
The third inequality comes from the estimate for the prediction on the integer line 
\(\E_{r\sim\mu}[\ell_\psi(Y,t)/B]\le h_i(t)\), and the median bound
gives \(\ell_\psi(x_{\mathrm{med}},t)/B\le 2h_i(t)\).

\paragraph{Use one fixed rate on the interval}

Similarly, suppose the current interval be $\{b+1,\cdots,k\}$. For each residue $r \in \{0,1,\cdots,K-1\}$, let $D_i(r)$ denote the cumulative
normalized loss through coordinate \(i\) obtained by using \(r\) as
the full residue at every coordinate from \(b+1\) to \(i\). Explicitly,
\[
D_i(r)
=\sum_{j=b+1}^{i}
  \frac{\ell_\psi\bigl(\phi_j(r-\tau_j(y_j)),y_j\bigr)}{B}.
\]
Initially, \(D_b(r)=0\) for every \(r\). After observing \(y_i\), we
update each residue by \(D_i(r)=D_{i-1}(r)+\frac{\ell_\psi(\phi_i(r-\tau_i(y_i)),y_i)}B\).
Thus, before predicting at coordinate \(i\), the scores \(D_{i-1}(r)\)
for all residues \(r\) are determined by the preceding labels.

For each \(i\in I\), define
\[
\mu_i(r)=\frac{e^{\beta D_{i-1}(r)}}
               {\sum_{s=0}^{K-1}e^{\beta D_{i-1}(s)}}.
\]
Apply the one-coordinate construction with the expert weights
\(p_{i,a}\) and the residue law \(\mu_i\) to obtain \(\phi_i\).
Using the observed labels, predict
\[
\widehat y_i=\phi_i\!\left(\sum_{j\ne i}\tau_j(y_j)\right).
\]
Let \(\Theta=[\sum_{j=1}^{n}\tau_j(y_j)]_K\). Since \(\phi_i\)
is \(K\)-periodic, \(\widehat y_i=\phi_i(\Theta-\tau_i(y_i))\).
Therefore,
\[
D_k(\Theta)=\sum_{j=b+1}^{k}
  \frac{\ell_\psi(\widehat y_j,y_j)}{B}.
\]

Applying the $\textbf{one-coordinate estimate}$
with \(\mu=\mu_i\) and \(t=y_i\), we obtain
\[
\E_{r\sim\mu_i}\bigl[D_i(r)-D_{i-1}(r)\bigr]
=
\E_{r\sim\mu_i}\!\left[
  \frac{\ell_\psi\bigl(\phi_i(r-\tau_i(y_i)),y_i\bigr)}{B}
\right]
\le (1+M^{-8})h_i(y_i).
\]
Each increment \(D_i(r)-D_{i-1}(r)\) lies in \([0,1]\), so we may use
the chord inequality \(e^{\beta u}\le 1+(e^\beta-1)u\) for
\(0\le u\le1\). The initial exponential sum equals \(K\), since
\(D_b(r)=0\) for every \(r\), and the final sum contains the term
\(e^{\beta D_k(\Theta)}\). Thus
\begin{align*}
\sum_{i=b+1}^{k}\frac{\ell_\psi(\widehat y_i,y_i)}{B}
&= D_k(\Theta)
 \le \frac{1}{\beta}\log\!\left(\sum_{r=0}^{K-1}e^{\beta D_k(r)}\right)\\
&= \frac{\log K}{\beta}
 + \frac{1}{\beta}\sum_{i=b+1}^{k}
   \log\!\left(
     \frac{\sum_{r=0}^{K-1}e^{\beta D_i(r)}}
          {\sum_{s=0}^{K-1}e^{\beta D_{i-1}(s)}}
   \right)\\
&= \frac{\log K}{\beta}
 + \frac{1}{\beta}\sum_{i=b+1}^{k}
   \log\!\left(
     \sum_{r=0}^{K-1}\mu_i(r)
     e^{\beta(D_i(r)-D_{i-1}(r))}
   \right)\\
&\le \frac{\log K}{\beta}
 + \frac{1}{\beta}\sum_{i=b+1}^{k}
   \log\!\left(
     1+(e^\beta-1)\E_{r\sim\mu_i}[D_i(r)-D_{i-1}(r)]
   \right)\\
&\le \frac{\log K}{\beta}
 + \frac{e^\beta-1}{\beta}(1+M^{-8})
   \sum_{i=b+1}^{k}h_i(y_i)\\
&\le \frac{\log K}{\beta}
 + \frac{e^\beta-1}{\beta}(1+M^{-8})(1+\beta)
   \left(L_I^*+\frac{\log M}{\beta}\right)\\
&\le \frac{\log K}{\beta}
 + (1+\beta)^3\left(L_I^*+\frac{\log M}{\beta}\right)\\
&\le L_I^*+7\beta L_I^*
 + \frac{8\log M+\log K}{\beta}\\
&\le L_I^*+14R\beta+\frac{19\log M}{\beta}
 = L_I^*+33\sqrt{R\log M}.
\end{align*}

The first two equalities after the initial bound follow by telescoping
and by the definition of \(\mu_i\), respectively. After applying the
chord inequality, we used \(\log(1+x)\le x\) and the one-coordinate
estimate above. Lemma~\ref{lem:mass}, applied to the best expert on
\(I\), then bounds \(\sum_{i=b+1}^{k}h_i(y_i)\) by
\((1+\beta)(L_I^*+(\log M)/\beta)\). For the numerical bounds, \(0<\beta\le1\) gives
\(e^\beta-1\le\beta+\beta^2\), and \(R<M^8\) gives
\(M^{-8}\le\sqrt{(\log M)/R}=\beta\). We used
\((1+\beta)^3\le1+7\beta\) for the coefficient of \(L_I^*\)
and \((1+\beta)^3\le8\) for the coefficient of \((\log M)/\beta\).
The last line follows from \(L_I^*\le R+1\le2R\),
\(\log K\le11\log M\), and
\(R\beta=(\log M)/\beta=\sqrt{R\log M}\).

\subsection{Large loss scales}

Next we will consider the large loss scales cases. Use the same notation above. Suppose $R \geq M^{8}$ and the interval is $I=\{b+1,\cdots,k\}$. 

At its start all \(M\) experts are present,
each has weight \(1/M\), and their cumulative costs are zero. Set
\[
  \beta=\sqrt{\frac{\log M}{R}},\qquad \eta=\frac{\beta}{8M}.
\]
Since \(M\ge2\), \(R\ge M^8\), and \(\log M\le M\), we have
\(R\ge256\), \(\beta\le M^{-7/2}<1/8\), and
\(R\beta=\sqrt{R\log M}\ge16\sqrt{\log 2}>4\).

\paragraph{Step one. Encode a label for the currently retained experts.}
Fix a coordinate \(i\), and let \(E\) be the set of experts present at
the start of round \(i\). For each possible label \(t\in[0,1]\), we
encode the vector of true normalized losses \((\ell_{i,a}(t))_{a\in E}\)
by rounding each entry upwards. Write \(c_{i,a}(t)\) for the rounded
cost and \(c_{i,a}=c_{i,a}(y_i)\) for the cost used in Lemma~3.1
once \(y_i\) has been observed.

If \(\max_{a\in E}\ell_{i,a}(t)\le 1/n\), assign \(t\) a single
distinguished symbol and set \(c_{i,a}(t)=1/n\) for every \(a\in E\).
Otherwise, let \(j\ge0\) be the unique integer for which
\[
  2^{-j-1}<\max_{a\in E}\ell_{i,a}(t)\le 2^{-j}.
\]
In this strip the grid spacing is \(\eta 2^{-j}\). The symbol records
\(j\) and the vector \((c_{i,a}(t))_{a\in E}\), where
\[
  c_{i,a}(t)=
  \min\!\left\{2^{-j},\
      \eta 2^{-j}\left\lceil
      \frac{\ell_{i,a}(t)}{\eta 2^{-j}}\right\rceil\right\}.
\]
Both entries in the minimum are at least \(\ell_{i,a}(t)\), and rounding
up adds at most one grid spacing. Thus in strip \(j\),
\[
  0\le c_{i,a}(t)-\ell_{i,a}(t)
  \le \eta 2^{-j}<2\eta\max_{a'\in E}\ell_{i,a'}(t).
\]
For the distinguished symbol the error is at most \(1/n\). In both cases
\(0\le\ell_{i,a}(t)\le c_{i,a}(t)\le1\), and
\begin{equation}
  0\le c_{i,a}(t)-\ell_{i,a}(t)
  \le 2\eta\max_{a'\in E}\ell_{i,a'}(t)+\frac1n.
  \label{eq:lg-one-round-cost}
\end{equation}

We also need a bound on the number of symbols in each strip. Fix \(j\)
and extend its rounding formula to all \(t\in[0,1]\). For each expert
\(a\), the function \(\ell_{i,a}(t)=\psi(|f^{(a)}(i)-t|)/B\) is
monotone on either side of \(f^{(a)}(i)\). On each side its rounded
cost is monotone and takes at most \(1+\lceil1/\eta\rceil\le3/\eta\)
distinct values. The preimage of each value on that side is an interval,
possibly a single point. Counting at most two endpoints per interval,
the experts together contribute at most
\(M\cdot2\cdot(3/\eta)\cdot2=12M/\eta\) endpoints. Between consecutive
endpoints every entry of the vector is constant. Counting the
endpoints themselves and the intervening intervals gives at most
\(2(12M/\eta)+1\le32M/\eta\) distinct vectors. The same bound holds
upon restriction to strip \(j\). Only strips with \(2^{-j}>1/n\) are
used, so the alphabet is finite. The encoding is Borel since it uses
comparisons and rounding of continuous loss functions.

After observing \(y_i\), add \(c_{i,a}\) to the cumulative cost of each
\(a\in E\). If \(\min_{a\in E}\sum_{q=b+1}^{i}c_{q,a}>R\), include
coordinate \(i\) and end the interval. Otherwise, remove every expert
whose cumulative cost exceeds \(2R\). At least one expert then has
cumulative cost at most \(R\) and remains. Call a maximal consecutive
part of \(I\) on which the set present at the start of each round stays
fixed a \emph{block}. A round that triggers a deletion is the last round
of its block. The surviving experts carry their cumulative costs into
the next block. Consequently there are at most \(M\) blocks in \(I\).
The last interval and its last block also end if the sample ends.

\paragraph{Step two. Control the cumulative rounding error.}
Fix a prefix \(\{b+1,\ldots,i\}\) of \(I\). For each expert \(a\),
consider its last round in this prefix at which it was present at the
start. Its cumulative cost before that round was at most \(2R\), and
the update in that round adds at most one. Thus its total true loss
over the rounds of the prefix in which it was present is at most \(2R+1\).
Bounding each round's maximum by the sum over the experts present in
that round gives
\[
  \sum_{q=b+1}^{i}
    \max_{\substack{a\ \text{present at the}\\\text{start of round }q}}
        \ell_{q,a}(y_q)
  \le \sum_{a=1}^{M}
       \sum_{\substack{b+1\le q\le i\\
                       a\ \text{present at the start of round }q}}
        \ell_{q,a}(y_q)
  \le M(2R+1).
\]
An expert \(a\) present at the start of round \(i\) has been present
since round \(b+1\). Summing \eqref{eq:lg-one-round-cost} through
round \(i\), including its update, therefore yields
\begin{equation}
\begin{aligned}
  0\le \sum_{q=b+1}^{i}\bigl(c_{q,a}-\ell_{q,a}(y_q)\bigr)
  &\le 2\eta\sum_{q=b+1}^{i}
    \max_{\substack{a'\ \text{present at the}\\\text{start of round }q}}
       \ell_{q,a'}(y_q)+\frac{i-b}{n}\\
  &\le 2\eta M(2R+1)+1
   =\frac{\beta}{4}(2R+1)+1\\
  &\le \frac34R\beta+1\le R\beta=\sqrt{R\log M}.
\end{aligned}
\label{eq:lg-prefix-error}
\end{equation}
The last line uses \(2R+1\le3R\) and \(R\beta\ge4\).

If \(a\) is removed after round \(i\), then its cumulative rounded cost satisfies \(\sum_{q=b+1}^{i} c_{q,a} >2R\). By \eqref{eq:lg-prefix-error}, its true loss on \(I\)
satisfies
\[
  \sum_{q=b+1}^{k}\ell_{q,a}(y_q)
  \ge\sum_{q=b+1}^{i}\ell_{q,a}(y_q)
  \ge\sum_{q=b+1}^{i}c_{q,a}-R\beta
  >2R-R\beta>R+1.
\]
At the start of the last round \(k\), an expert of smallest cumulative
cost has cost at most \(R\), by the stopping rule. Its cost after the
last update is at most \(R+1<2R\), so it survives and its true loss on
\(I\) is at most \(R+1\). It follows that a minimizer \(a^*\) of
the true loss defining \(L_I^*\) survives the entire interval.
Applying \eqref{eq:lg-prefix-error} to this expert at \(i=k\) gives
\begin{equation}
  L_I^*\le R+1,\qquad
  \sum_{i=b+1}^{k}c_{i,a^*}\le L_I^*+R\beta.
  \label{eq:lg-surviving-expert}
\end{equation}
If the stopping threshold is crossed, the final rounded cost of every
surviving expert exceeds \(R\). In particular,
\begin{equation}
  R<\sum_{i=b+1}^{k}c_{i,a^*}\le L_I^*+R\beta,
  \qquad L_I^*>R(1-\beta)>R/2.
  \label{eq:lg-crossing}
\end{equation}

\paragraph{Step three. Choose one expert for each observed sequence.}
Fix a block starting at coordinate \(l+1\), its retained set \(E\), and
the cumulative costs of its experts through round \(l\). For every
coordinate from \(l+1\) to \(n\), use the alphabet defined in step one
with this same \(E\). A row \(z=(z_{l+1},\ldots,z_n)\) in Lemma~4.1 is
a complete sequence of symbols from these alphabets. Starting with the
fixed cumulative costs, apply the stopping and deletion rules along
the row until the first round that ends the scale or removes an expert,
or until \(n\). Denote the resulting block, including its final
coordinate, by \(J(z)\). Whether \(i\in J(z)\) is determined by
\(z_{l+1},\ldots,z_{i-1}\).

For each \(l<i\le n\) and each sequence \(s\) with coordinate \(i\) omitted,
form the group \(g=(i,s)\). Its actions are the experts \(a\in E\).
Write \(c_{i,a}(z_i)\) for the rounded cost read from symbol \(z_i\).
For the actual symbol of \(y_i\), this equals \(c_{i,a}(y_i)\).
Set
\[
  c_{zga}
  =\mathbf{1}_{\{z_{-i}=s\}}\mathbf{1}_{\{i\in J(z)\}}c_{i,a}(z_i),
  \qquad g=(i,s),\quad a\in E.
\]
For a group with \(i\in J(z)\) and \(z_{-i}=s\), use the exponential
weights \(p_{ga}=p_{i,a}\) from Lemma~3.1. On the actual data these are
\[
  p_{i,a}=
  \frac{e^{-\beta\sum_{q=b+1}^{i-1}c_{q,a}}}
       {\sum_{a'\in E}e^{-\beta\sum_{q=b+1}^{i-1}c_{q,a'}}},
  \qquad
  h_i=\sum_{a\in E}p_{i,a}c_{i,a}.
\]
On row \(z\), the cumulative cost in the exponent is the entering
cost \(\sum_{q=b+1}^{l}c_{q,a}\) plus
\(\sum_{q=l+1}^{i-1}c_{q,a}(z_q)\). Thus all rows with \(z_{-i}=s\)
give the same weights and the same decision on including \(i\). For a
group whose coordinate is excluded all coefficients vanish, so assign
weight one to the smallest index in \(E\).

To apply Lemma~4.1, fix \(g=(i,s)\) and two actions \(a,a'\in E\).
Only rows with \(z_{-i}=s\) can contribute to the difference of their
columns. There is exactly one such row for each possible symbol \(z_i\).
The distinguished symbol contributes zero, since both costs then equal
\(1/n\). In strip \(j\) both costs lie in \([0,2^{-j}]\). Using the
symbol count from step one, we obtain
\[
  \sum_z(c_{zga}-c_{zga'})^2
  \le \sum_{j\ge0}\frac{32M}{\eta}(2^{-j})^2
  =\frac{128M}{3\eta}
  \le \left(7\sqrt{\frac{M}{\eta}}\right)^2.
\]
Take \(D_1=7\sqrt{M/\eta}\). Lemma~4.1 gives one choice
\(a_g\in\operatorname{supp}(p_g)\) for every group such that the total
rounding error is at most \(36D_1\) on every row simultaneously.

For the observed data, let \(z\) be the corresponding row and
\(J=J(z)\). At \(i\in J\), the predictor encodes all labels after
coordinate \(l\) except \(y_i\), using the fixed set \(E\). It chooses
\(a_i(y_{-i})=a_{(i,z_{-i})}\) and predicts
\(\widehat y_i=f^{(a_i(y_{-i}))}(i)\). On row \(z\), exactly the
groups \((i,z_{-i})\) with \(i\in J\) enter the block loss. Substituting
their costs in the conclusion of Lemma~4.1 gives
\begin{align*}
  \sum_{i\in J}\frac{\ell_\psi(\widehat y_i,y_i)}{B}
  &\le \sum_{i\in J} c_{i,a_{(i,z_{-i})}}(y_i)
   =\sum_g c_{z,g,a_g}\\
  &\le \sum_g\sum_{a\in E}p_{ga}c_{zga}+36D_1
   =\sum_{i\in J}h_i+36D_1.
\end{align*}
Summing over at most \(M\) blocks adds at most \(36MD_1\) to the
weighted loss on \(I\). Since \(\eta=\beta/(8M)\) and
\(\beta=\sqrt{(\log M)/R}\), this error satisfies
\begin{align*}
  36MD_1
  &=252M\sqrt{\frac{M}{\eta}}
   =252\sqrt{8}\,\frac{M^2}{\sqrt{\beta}}\\
  &=252\sqrt{8}\,
    \frac{M^2}{R^{1/4}(\log M)^{3/4}}\sqrt{R\log M}
   \le 2^{11}\sqrt{R\log M}.
\end{align*}
Here the large-scale assumption \(R\ge M^8\) gives \(M^2\le R^{1/4}\),
and \(M\ge2\) gives \((\log M)^{-3/4}<2\).

\paragraph{Step four. Compare with the best expert on the interval.}
Each surviving expert keeps its cumulative cost across block boundaries,
so its unnormalized weight is multiplied by \(e^{-\beta c_{i,a}}\)
in every round. Hence Lemma~3.1 applies once on the whole interval \(I\).
Use the surviving
minimizer \(a^*\) from \eqref{eq:lg-surviving-expert} and the
identity \((\log M)/\beta=R\beta\). Then
\begin{align*}
  \sum_{i=b+1}^{k}h_i
  &\le (1+\beta)
       \left(\sum_{i=b+1}^{k}c_{i,a^*}+\frac{\log M}{\beta}\right)
   \le (1+\beta)(L_I^*+2R\beta)\\
  &=L_I^*+2R\beta+\beta(L_I^*+2R\beta)
   \le L_I^*+4R\beta.
\end{align*}
The last inequality follows from
\(L_I^*+2R\beta\le R+1+2R\beta\le2R\).
Adding the rounding error from step three yields
\begin{equation}
\begin{aligned}
  \sum_{i=b+1}^{k}\frac{\ell_\psi(\widehat y_i,y_i)}{B}
  &\le \sum_{i=b+1}^{k}h_i+36MD_1\\
  &\le L_I^*+(4+2^{11})\sqrt{R\log M}
   \le L_I^*+2^{12}\sqrt{R\log M}.
\end{aligned}
\label{eq:lg-interval-loss}
\end{equation}

\subsection{Complete proof}

Fix a complete sample \(S\) and let \(R_{\mathrm{last}}\) be the scale
of its final interval. If there is just one interval, then
\(R_{\mathrm{last}}=\log(2M)\). Otherwise let \(I\) be the interval
immediately preceding the last one. Its scale was \(R_{\mathrm{last}}/4\),
and it crossed its stopping threshold. The stopping rule in Section~5.1
gives \(L_I^*>R\) at a small-scale crossing, while
\eqref{eq:lg-crossing} gives \(L_I^*>R/2\) at a large-scale
crossing. Thus
\[
  L_n^*\ge L_I^*>\frac12\,\frac{R_{\mathrm{last}}}{4}
  =\frac{R_{\mathrm{last}}}{8}.
\]
The first inequality holds because the loss of a full-sample minimizer
on \(I\) is at least \(L_I^*\) and at most its loss on all \(n\)
coordinates. In both cases \(R_{\mathrm{last}}\le\log(2M)+8L_n^*\).
Since each restart multiplies \(R\) by four, summing backwards from the
last interval gives
\[
  \sum_{\text{scales }R}\sqrt{R\log M}
  \le\sqrt{R_{\mathrm{last}}\log M}\sum_{j=0}^{\infty}2^{-j}
  =2\sqrt{R_{\mathrm{last}}\log M}.
\]

Choose an expert \(a^*\) attaining \(L_n^*\) on this sample.
For each interval \(I\), its loss there is an upper bound for
\(L_I^*\). Hence
\[
  \sum_I L_I^*
  \le \sum_I\sum_{i\in I}\ell_{i,a^*}(y_i)
  =\sum_{i=1}^n\ell_{i,a^*}(y_i)=L_n^*.
\]
Section~5.1 bounds the small-scale excess by \(33\sqrt{R\log M}\),
and \eqref{eq:lg-interval-loss} applies at every large scale. Since
\(33\le2^{12}\), summing these bounds gives
\begin{align*}
  \sum_{i=1}^{n}\frac{\ell_\psi(\widehat y_i,y_i)}{B}
  &\le \sum_I L_I^*
      +2^{12}\sum_{\text{scales }R}\sqrt{R\log M}\\
  &\le L_n^*+2^{13}\sqrt{R_{\mathrm{last}}\log M}
   \le L_n^*+2^{13}\sqrt{\bigl(\log(2M)+8L_n^*\bigr)\log M}\\
  &\le L_n^*+2^{15}\left(\sqrt{L_n^*\log M}+\log M\right).
\end{align*}
In the last line we used \(\log(2M)\le2\log M\) and
\(\sqrt{u+v}\le\sqrt u+\sqrt v\) for \(u,v\ge0\).
By the normalization in (5.1),
\[
  \frac{B}{n}L_n^*=\widehat L^*_{S,\psi},\qquad
  \frac{B}{n}\sqrt{L_n^*\log M}
  =\sqrt{\frac{B\widehat L^*_{S,\psi}\log M}{n}}.
\]
Multiplying the total-loss bound by \(B/n\) therefore gives
\[
  \widehat L^{\mathrm{LOO}}_{S,\psi}(A)
  \le \widehat L^*_{S,\psi}
   +2^{15}\left(
       \sqrt{\frac{B\widehat L^*_{S,\psi}\log M}{n}}
       +\frac{B\log M}{n}\right),
\]
which implies the inequality in Theorem \ref{thm:GEN}.

\paragraph{One deterministic rule for all samples.}
At the end of this section, we will give the whole deterministic rule of Theorem \ref{thm:GEN}. 

At coordinate \(i\), the labels \(y_1,\ldots,y_{i-1}\) determine the
current scale and its starting coordinate. At a small scale they also
determine the expert weights and the map \(\phi_i\), since the scores
of all residues are updated after each preceding round. The predictor
computes \(\phi_i(\sum_{j\ne i}\tau_j(y_j))\) and chooses the smallest
expert index giving that value at \(i\).

At a large scale the same prefix determines the current block, its
retained set, and its entering costs. All coordinates in the block
therefore recover the same table once the choice of a feasible table is
fixed. To make this choice deterministic, order the groups and the
expert indices in advance. Among the tables supported by the group
weights, choose the lexicographically first table satisfying
\[
  \left|\sum_g c_{z,g,a_g}-\sum_g\sum_{a\in E}p_{ga}c_{zga}\right|
  \le36D_1
  \qquad\text{for every row }z.
\]
Lemma~4.1 guarantees such a table. The predictor reads its expert from
the entry \((i,z_{-i})\), which is determined by \(y_{-i}\).

The small-scale ranks, median, origin, and integer thresholds are
determined by finite comparisons of Borel functions. At a large scale,
for a fixed block start, \(E\), and \(R\), the alphabets, rows, and
candidate tables are finite. For each table its support conditions and
its row inequalities are Borel in the entering costs, since each row
uses finitely many updates and comparisons. Choosing the first feasible
table is therefore Borel. Together these constructions define
deterministic Borel maps \(a_i(y_{-i})\in[M]\), with
\(\widehat y_i=A_i(y_{-i})=f^{(a_i(y_{-i}))}(i)\), and the bound
holds for every sample \(S\).

\section{Discussions}

In this section, we will first discuss that our theorem is minimax optimal up to constants. To begin with, we restate our theorem:

Fix $n,M\ge2$ and an admissible $\psi$, with $B=\psi(1)>0$.
For every $0\le r\le B$,
\[
\begin{aligned}
 &\sup_{\substack{\mathcal F\subset[0,1]^{[n]}\\ |\mathcal F|=M}}
   \ \inf_{A\ \mathrm{deterministic,\ proper}}
   \ \sup_{\substack{v_y\in[0,1]^n\\
                      \widehat L^*_{S,\psi}\le r}}
   \left(\widehat L^{\mathrm{LOO}}_{S,\psi}(A)
                  -\widehat L^*_{S,\psi}\right)\asymp
    \min\!\left\{B,\
      \sqrt{\frac{B r\log M}{n}}+\frac{B\log M}{n}\right\}.
\end{aligned}
\]
The constants are universal. The infimum is over the Borel prediction
maps of Section~2.1 for the given trace $\mathcal F$. The upper bound
is attained up to a universal factor by one rule for all $r$. The
lower bound continues to hold if the rule knows $r$ and may randomize,
with its loss then taken in expectation on each fixed sample.

\begin{proof}
For the upper bound, it is directly shown by applying Theorem~\ref{thm:GEN} to each sample satisfying
$\widehat L^*_{S,\psi}\le r$. Since each loss is at most $B$ and the
oracle loss is nonnegative, the excess is also at most $B$.

\providecommand{\E}{\mathbb{E}}
It is sufficient to prove the lower bound. We construct a trace
$\mathcal F$ of size $M$ and a distribution on samples for which the
best expert has loss at most $r$ on every realization.

\medskip\noindent
\textbf{Step one. Construct the trace.}
Reserve coordinate $n$ and take $q$ blocks of length $w$, where
\[
 q=\min\{n-1,\lfloor\log_2 M\rfloor\},\qquad
 w=\max\!\left\{1,\left\lfloor\frac{(n-1)r}{Bq}\right\rfloor\right\}.
\]
These blocks fit in $[n-1]$. Indeed, $qw=q\le n-1$ if $w=1$, and
$qw\le(n-1)r/B\le n-1$ if $w\ge2$. Divide the first $qw$ coordinates
into consecutive blocks of length $w$. For each of the $2^q$ ways of
choosing one bit per block, include the expert that predicts each
block's chosen bit throughout that block and zero elsewhere. If $2^q<M$,
add $M-2^q$ experts that predict zero on $[n-1]$ and take distinct
values in $(0,1]$ at coordinate $n$. Thus $|\mathcal F|=M$, and the
available predictions at each of the first $qw$ coordinates are $0$ and $1$.

Let $\epsilon_1,\ldots,\epsilon_{qw}$ be independent Rademacher signs.
Set $y_i=(1+\epsilon_i)/2$ on the first $qw$ coordinates and $y_i=0$
elsewhere. A best expert can choose a most frequent bit on each block
and predict zero elsewhere. The added experts agree with the all-zero
expert on $[n-1]$ and incur nonnegative loss at $n$. Thus the best
expert's total loss is exactly $B$ times the sum, over the blocks, of
the smaller of the numbers of zeros and ones. For $w=1$ this loss is zero.
For $w\ge2$, every sample
satisfies
\[
 \widehat L^*_{S,\psi}
 \le \frac{Bq}{n}\left\lfloor\frac{w}{2}\right\rfloor
 \le \frac{Bqw}{2n}
 \le \frac{(n-1)r}{2n}
 \le r.
\]

\medskip\noindent
\textbf{Step two. Compare the rule with the best expert.}
Fix any proper rule $A$, which may know $r$ and use independent
randomness. At each of the first $qw$ coordinates, condition on $y_{-i}$
and the rule's random choices. Then $\widehat y_i=A_i(y_{-i})$ is a
fixed bit and $y_i$ is still a fair bit. A mistake costs $B$ and a
correct prediction costs zero, so the rule's expected loss at this
coordinate is $B/2$.

On the first block let $Z=\sum_{i=1}^w\epsilon_i$. The numbers of ones
and zeros are $(w+Z)/2$ and $(w-Z)/2$, so their minimum is
$(w-|Z|)/2$. Since every block has the same distribution, the best
expert makes $(w-\E|Z|)/2$ mistakes per block in expectation. Hence,
taking expectation over the labels and any randomness of the rule,
\[
\begin{aligned}
 \E\!\left[\widehat L^{\mathrm{LOO}}_{S,\psi}(A)
                -\widehat L^*_{S,\psi}\right]
 &\ge \frac{Bqw}{2n}-\frac{Bq}{2n}\bigl(w-\E|Z|\bigr)
  = \frac{Bq}{2n}\,\E|Z|.
\end{aligned}
\]
Independence and symmetry give $\E Z^2=w$ and
$\E Z^4=w+6\binom{w}{2}=3w^2-2w$, since only products in which each
sign occurs an even number of times contribute. By H\"older's inequality,
\[
 \E Z^2
 =\E\!\left[|Z|^{2/3}|Z|^{4/3}\right]
 \le (\E|Z|)^{2/3}(\E Z^4)^{1/3}.
\]
It follows that
\[
 \E\!\left[\widehat L^{\mathrm{LOO}}_{S,\psi}(A)
                -\widehat L^*_{S,\psi}\right]
 \ge \frac{Bq}{2n}\,\frac{(\E Z^2)^{3/2}}{(\E Z^4)^{1/2}}
 = \frac{Bq}{2n}\,\frac{w^{3/2}}{\sqrt{3w^2-2w}}
 \ge \frac{Bq\sqrt{w}}{2\sqrt{3}\,n}.
\]

\medskip\noindent
\textbf{Step three. Conclude the minimax bound.}
The inequalities $n-1\ge n/2$ and
$\max\{1,\lfloor x\rfloor\}\ge x/2$ for $x\ge0$ give
\[
 q\ge\tfrac12\min\{n,\log_2 M\}\ge\tfrac12\min\{n,\log M\},
 \qquad
 w\ge\max\!\left\{1,\frac{nr}{4Bq}\right\}.
\]
If $\log M\le n$, these estimates yield
\[
\begin{aligned}
 \frac{Bq\sqrt w}{2\sqrt{3}\,n}
 &\ge \max\!\left\{\frac{Bq}{2\sqrt{3}\,n},
              \frac{1}{4\sqrt{3}}\sqrt{\frac{Brq}{n}}\right\}\\
 &\ge \max\!\left\{\frac{B\log M}{4\sqrt{3}\,n},
              \frac{1}{4\sqrt{6}}\sqrt{\frac{Br\log M}{n}}\right\}\\
 &\ge \frac{1}{8\sqrt{6}}
       \left(\sqrt{\frac{Br\log M}{n}}+\frac{B\log M}{n}\right).
\end{aligned}
\]
If $\log M>n$, then $q\ge n/2$ and $w\ge1$, giving the lower bound
$B/(4\sqrt{3})$. In this case the minimum in Theorem~\ref{minimax} equals $B$.

For each fixed rule $A$, some sample in the support has expected excess
loss at least as large as the average in Step two. On this fixed sample
the expectation is over the rule's randomness alone. By Step one the
sample satisfies $\widehat L^*_{S,\psi}\le r$, and the same trace
works for every $A$. Taking the infimum over $A$ and then the supremum
over traces of size $M$ proves the lower bound with constant
$1/(8\sqrt{6})$.

\end{proof}

In particular, Theorem~\ref{thm:GEN} is minimax optimal up to universal constants
over samples with $\widehat L^*_{S,\psi}\le r$, after taking the
trivial cap $B$ into account. At $r=0$ the rate is
$B\min\{1,(\log M)/n\}$. At $r=B$ the budget restriction disappears
and the worst-case minimax rate is $B\min\{1,\sqrt{(\log M)/n}\}$.

\vspace{1em}

What is more, we can give two conjectures based on our results. The first conjecture is given by removing the coefficient before $\sqrt{\frac{B\widehat L^*_{S,\psi}\log M}{n}}$. We ask whether the square root coefficient can equal one. The second conjecture is given because of the possibility that more localized variable like Vapnik-Chervonenkis (VC) dimension can provide the bound as well. It is inspired by the result of Mathiasen, Qian, and Zhivotovskiy~\cite[Theorem~2]{mathiasen2026optimal}

\paragraph{Conjecture 1 (Unit leading constant).}
Fix $n,M\ge2$, a trace $\mathcal F\subseteq[0,1]^{[n]}$ of size $M$,
and an admissible $\psi$ with $B=\psi(1)>0$. There is a deterministic
proper leave-one-out rule $A$, depending only on $n,\mathcal F,\psi$,
such that for every sample $S$,
\[
 \widehat L^{\mathrm{LOO}}_{S,\psi}(A)-\widehat L^*_{S,\psi}
 \le
 \sqrt{\frac{B\widehat L^*_{S,\psi}\log M}{n}}
 + C\frac{B\log M}{n},
\]
where $C$ is universal. The coefficient of the square root is one.

\paragraph{Conjecture 2 (A pseudo-dimension bound).}
Under the same assumptions, let $d$ be the pseudo-dimension of
$\mathcal F$, namely the VC dimension of its subgraph class
\[
 \bigl\{(i,t)\mapsto\mathbf 1\{f(i)>t\}: f\in\mathcal F\bigr\},
 \qquad (i,t)\in[n]\times[0,1].
\]
There is a deterministic proper leave-one-out rule $A$, depending only on
$n,\mathcal F,\psi$, such that for every sample $S$,
\[
 \widehat L^{\mathrm{LOO}}_{S,\psi}(A)-\widehat L^*_{S,\psi}
 \le C\left(
       \sqrt{\frac{B\widehat L^*_{S,\psi}\,d}{n}}
       +\frac{Bd}{n}\right),
\]
where $C$ is universal. In both conjectures the predictions
$A_i(y_{-i})$ are Borel functions taking values in
$\{f(i):f\in\mathcal F\}$.

\section*{AI use disclosure}

The original proof of Theorem \ref{thm:GEN} was discovered by GPT-6-Astra, GPT-6.2-Sol, and GPT-6.2-Astra. The authors checked the original proof, then simplified the proofs and revised the arguments. The authors take full responsibility for their mathematical content, and
independently verified the extensions presented here.

\end{document}